\documentclass[runningheads]{llncs}
\usepackage{graphicx}
\usepackage[misc]{ifsym}
\usepackage{multirow}
\usepackage{booktabs}
\usepackage{arydshln}
\usepackage{amsfonts}       
\usepackage{amsmath}
\usepackage{bbm}
\usepackage{color}
\usepackage[pagebackref=true,breaklinks=true,colorlinks=true,bookmarks=false,citecolor=blue,urlcolor=blue,linkcolor=blue]{hyperref}

\renewcommand{\b}{\boldsymbol}
\newcommand{\U}{\mathcal{U}}
\newcommand{\B}{\mathcal{B}}
\newcommand{\R}{\mathbb{R}}
\newcommand{\cL}{\mathcal{L}}
\begin{document}
\title{Dynamic Bank Learning for Semi-supervised Federated Image Diagnosis with Class Imbalance}
\titlerunning{Semi-supervised Federated Image Diagnosis with Class Imbalance}
%
\author{Meirui Jiang\inst{1} \and
Hongzheng Yang\inst{2} \and
Xiaoxiao Li\inst{3} \and 
Quande Liu\inst{1} \and \\
Pheng-Ann Heng\inst{1} \and
Qi Dou\textsuperscript{1(\Letter)}
}
\authorrunning{M. Jiang et al.}
\institute{Dept. of Computer Science and Engineering, The Chinese University of Hong Kong \\
\and Dept. of Artificial Intelligence, Beihang University 
\and Dept. of Electrical and Computer Engineering, The University of British Columbia 
}
\maketitle              
\begin{abstract}
Despite recent progress on semi-supervised federated learning (FL) for medical image diagnosis, the problem of imbalanced class distributions among unlabeled clients is still unsolved for real-world use. In this paper, we study a practical yet challenging problem of class imbalanced semi-supervised FL (imFed-Semi), which allows all clients to have only unlabeled data while the server just has a small amount of labeled data. This imFed-Semi problem is addressed by a novel dynamic bank learning scheme, which improves client training by exploiting class proportion information. This scheme consists of two parts, i.e., the dynamic bank construction to distill various class proportions for each local client, and the sub-bank classification to impose the local model to learn different class proportions. We evaluate our approach on two public real-world medical datasets, including the intracranial hemorrhage diagnosis with 25,000 CT slices and skin lesion diagnosis with 10,015 dermoscopy images. The effectiveness of our method has been validated with significant performance improvements (7.61\% and 4.69\%) compared with the second-best on the accuracy, as well as comprehensive analytical studies.
Code is available at \url{https://github.com/med-air/imFedSemi}.


\keywords{Semi-supervised FL  \and Class Imbalance \and Image Diagnosis.}
\end{abstract}

\section{Introduction}
Federated learning (FL), i.e., collaboratively training a model with decentralized datasets, is increasingly important for medical image diagnosis~\cite{dong2021federated,dou2021federated,li2021fedbn,roth2020federated,sheller2020federated,wu2021federated}. To significantly enlarge cohorts, developing semi-supervised FL is crucial to include massive unlabeled data from different training clients~\cite{fedmatch,semifl_covid,zhang2020benchmarking}. Though some progress on semi-FL has been made for image diagnosis tasks~\cite{bdair2021fedperl,fedirm}, there are still key challenges remaining to be solved. First, existing methods typically assume one or several client(s) to be fully labeled, without allowing the extreme yet practically favorable situation that \emph{all} clients are unlabeled. Second, \emph{class imbalance} issue among clients (due to different patient demographics and disease incidence rate~\cite{rieke2020future,sheller2020federated}) hinders model accuracy, yet not been carefully considered.
\\\indent In this regard, we address a new problem setting of class imbalanced semi-supervised FL (named as \emph{imFed-Semi}), in which we assume that only the server holds a small amount of labeled data, while all clients only provide unlabeled data with the presence of class imbalance. This is a difficult problem, but closer to real-world use compared with previous ideal ones. Existing semi-supervised methods typically apply the consistency regularization~\cite{gyawali2020semi,fedirm,shi2020graph,sohn2020fixmatch} or use the pseudo-labeling~\cite{bai2017semi,bdair2021fedperl,semifl_covid,zhang2020benchmarking} on unlabeled samples. For instance, FedPerl~\cite{bdair2021fedperl} obtains pseudo labels for unlabeled data by ensembling models for a cluster of similar clients, and FedIRM~\cite{fedirm} imposes the consistency regularization by adding a class relation constraint between labeled and unlabeled clients. Unfortunately, these methods require some clients to be partially or fully labeled. Moreover, they suffer from performance drop upon the class imbalance problem, i.e., unlabeled clients have different proportions of samples from each disease category. Such a limitation is because that, the sample-level supervision on unlabeled data leads to each client model training to be locally dominated by its own majority class(es), therefore affecting the model aggregation at the server in FL.
\\\indent The key point to address the imFed-Semi lies in how to design class proportion-aware supervision to enhance training for unlabeled clients. Our insight is to rely on labeled data at the server to help unlabeled clients for estimating class-specific proportions.
Inspired by the learning from label proportion works~\cite{dulac2019deep,lu2022unsupervised,quadrianto2009estimating}, we can arrange data into several subsets and leverages the information of label proportion of subsets to weakly supervise the unlabeled client model training. In other words,
we can split data inside the client to distill various label proportions from pseudo labels, because it makes more sense to use the soft global information instead of unreliable sample-level pseudo labels to train the model, especially when all samples in the client are unlabeled. 
Then, how to obtain an accurate proportion estimation becomes an important step, directly using pseudo labels is unreliable due to misclassification issues on hard samples.
\\\indent In this paper, we propose a novel dynamic bank learning method for the imFed-Semi problem. Our method consists of two parts, the dynamic bank construction to extract various class proportion information within each client, and the sub-bank classification to enforce the local model to learn different class proportions. Specifically, the dynamic bank iteratively collects highly-confident samples during the training to estimate the client class distribution, and splits samples into sub-banks with the presence of different pseudo label proportions. Furthermore, a prior transition function is designed to transform the original classification task into the sub-bank classification task, which explicitly utilizes different class proportions to train the local model.
Owing to such label-proportion-awared supervision, the local client training is enhanced to learn different distributions of imbalanced classes to avoid being dominated by the local majority class(es).
We validate our method on two large-scale real-world medical datasets, including the intracranial hemorrhage diagnosis with 25,000 CT slices and the skin lesion diagnosis with 10,015 dermoscopy images. The effectiveness of our method has been validated with significant performance improvements on both tasks compared with a number of state-of-the-art semi-supervised learning and FL methods, as well as comprehensive analytical studies.

\section{Method}
\begin{figure}[t!]
\centering
\includegraphics[width=0.99\textwidth]{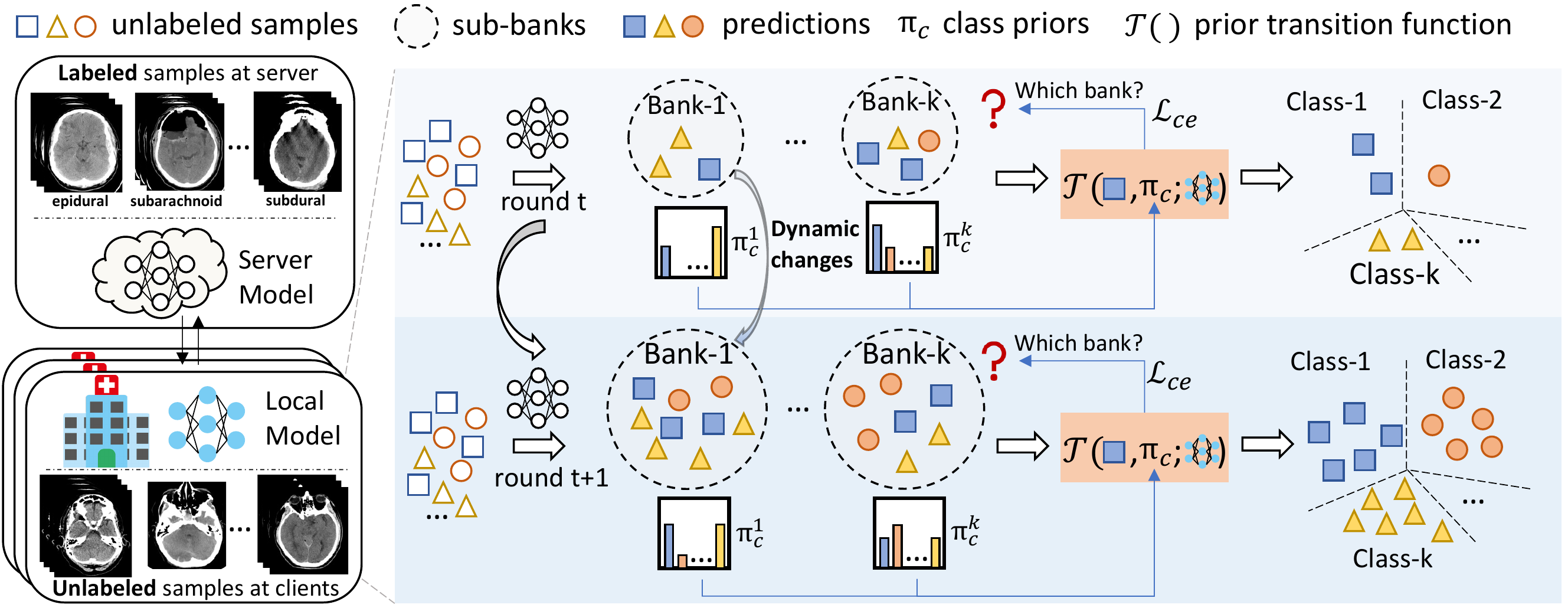}
\caption{Method overview. Our approach constructs sub-banks to collect confident samples and estimate class priors ($\pi_c^k$) for each client $c$. Then a sub-bank classification task is designed via learning different class proportions using the prior transition function. Finally, local model is promoted to learn discriminative decision boundaries.}
\label{fig:overview}
\end{figure}
\subsection{Overview of the imFed-Semi Framework}
Hereby, we formulate the semi-supervised federated classification task with class imbalance.
Fig.~\ref{fig:overview} is an overview of our proposed imFed-Semi solution.
We consider a K-class classification problem with input space $\mathcal{X} \subset \R^d$ and label space $\mathcal{Y}$, where $d$ is the input dimension.
Let $\b x \in \mathcal{X}$ and $y\in\mathcal{Y}$ be the input and output following an underlying joint distribution with density $p(\b x,y)$, which is identified by the class priors
$\{\pi^{k}=p(y=k)\}_{k=1}^{K}$ and the class-conditional densities $\{p(\b x \mid y=k)\}_{k=1}^{K}$. 
We have $n_s$ labeled samples $\mathcal{S}=\{(\b x^s_i,y^s_i)\}_{i=1}^{n_s}$ at server, and for $C$ clients, each client $c \in [C]$ has $n_c~(n_c \! \gg \! n_s)$ unlabeled samples $\U_c = \{(\b x^c_i)\}_{i=1}^{n_c}$. 
The goal of the classification task is to obtain a global FL model $f_g:\mathcal{X} \rightarrow \mathcal{Y}$ by using several labeled samples at the server and large-scale unlabeled data with the presence of such class-imbalanced clients. 

The FL training paradigm of our framework adopts the popular FedAvg~\cite{fedavg} algorithm, where the server collects client models $f_c$ and aggregates them to obtain the global model $f_g = \sum_{c=1}^{C}\frac{n_c}{N}f_c$, where $N=\sum_{c=1}^C n_c$. 
Then, the global model is broadcast back to clients for local training, and such a process repeats until convergence. The special practice in our label-at-server setting is, before broadcasting, the global model is updated for an extra round of gradient descents with labeled samples at the server.
Overall, the FL loss function is written as:
\begin{equation}
\label{eq:loss_all}
\cL_{overall} = \cL_s (f_g(\b x^s),y^s) + \frac{1}{C} \sum\nolimits_{c=1}^{C}\cL_u(f_g(\b x^c)),
\end{equation}
where the global model $f_g$ performs supervised learning on labeled sample pairs $\mathcal{S}=\{(\b x^s_i,y^s_i)\}_{i=1}^{n_s}$ by minimizing the cross-entropy loss $\cL_s$ at server. The local clients perform unsupervised learning which minimizes $\cL_u$ (cf. Eq.~(\ref{eq:L_u})) on unlabeled inputs $\U_c = \{(\b x^c_i)\}_{i=1}^{n_c}$ with our designed new scheme in the following.

\subsection{Dynamic Bank Construction for Unlabeled Clients}
Considering current semi-supervised methods are prone to make the local training dominated by majority class(es) at each client, we instead encourage the FL model to learn discriminative decision boundaries by taking advantage of different class proportions from a global perspective. Inspired by learning from class proportions of unlabeled samples~\cite{lu2022unsupervised}, we design a \textit{dynamic bank learning} scheme, which estimates the class proportions for client training. 
Specifically, we first build a dynamic bank to collect confident samples based on the thresholding of high prediction probabilities. Then, we split the dynamic bank into $K$ sub-banks to present different class proportions locally. By further designing a class prior transition function, we convert the $K$-class classification problem into $K$-bank classification via learning different class proportions.
\\
\\
\textbf{Dynamic bank construction.}
We build the dynamic bank by progressively storing confident samples as training goes on. 
Given an unlabeled sample $\b x_i^c$ at each client, we denote $p_i$ ($c$ is omitted for ease of notation) as the highest prediction probability across all classes by the local model, i.e., $p_i = \max{f_c(\b{x_i^c})}$.
We use a threshold $\tau_\alpha$ to select confident samples exceeding a high probability. In addition, 
due to class imbalance, the minority class(es) tend to be underrepresented~\cite{shen2022an}, leading to lower prediction probability by the local model. We further use another threshold $\tau_\beta$ to rescue underrepresented unlabeled samples, which
helps maintain the class diversity. With these considerations, we design the following threshold scheme to dynamically collect samples:
\begin{equation}
\label{eq:bank}
\B_{c,t} = \B_{c,t-1} \setminus \{\mathbbm{1}_{(p_i^{t-1}<\tau_\alpha)} \cdot \b x_i^c\}_{i=1}^{n_c} \cup \{\mathbbm{1}_{(p_i^t>\tau_\beta)} \cdot \b x_i^c\}_{i=1}^{n_c}.
\end{equation}
Specifically, $\B_{c,t}$ is the bank for client $c$ at the $t$-th communication round and $p_i^t$ is predicted probability at the round $t$.
The bank is initialized as an empty set at the beginning, i.e., $\B_{c,0} = \phi$. Before each round, the bank collects samples with probability exceeding the threshold $\tau_\beta$ for training. Then it is adjusted by only preserving confident samples with prediction probability exceeding the $\tau_\alpha$. By building the bank in such a dynamic way, we can gradually gather and make use of more samples at each round to estimate class distributions. In particular, almost all samples are promised to be included in the bank as training goes on, which helps the local model to get rid of neglecting minor classes. 
\\
\\
\textbf{Class prior estimation.}
We further use the proportion of classes in the constructed dynamic bank to provide supervision for the training of unlabeled clients.
The idea is to emphasize the class prior knowledge by proposing an auxiliary task of sub-bank classification, which explicitly connects the original classification task with class priors. The ground-truth for the sub-bank classification task can be easily obtained during training, thus providing more reliable supervision than conventional pseudo-labeling.
In practice, for each bank $\B_c$ of client $c$, we randomly split it to $K$ non-overlapping sub-banks, i.e., $\B_c = \{\B_c^m\}_{m=1}^K$, where $\B_c^m = \{\b x_i^{c,m}\}_{i=1}^{n_{c,m}}$ is the $m$-th sub-bank with sample size $n_{c,m}$. Notably, we impose that not all sub-banks have exactly the same class distribution,
considering diverse training cases for real-world applications. Denoting $\bar{y}$ as the index of these $K$ sub-banks, we utilize the index as proxy labels to transform unlabeled samples to labeled pairs $\B_c = \{(\b x^c_i,\bar{y}_i)\}_{i=1}^{n_b} \sim \bar{p}_c(\b x,\bar{y})$, where $n_b$ is the bank size and $\bar{p}_c(\b x,\bar{y})$ is the underlying joint distribution for the random variables $\b x\in \mathcal{X}$ and $\bar{y} \in [K]$. Since we assume the original input and output follow the distribution $p(\b x,y)$, which can be identified by class priors and class-conditional densities, the data samples of each sub-bank can be seen as drawn from a mixture of original class-conditional densities as follows:
\begin{equation}
\label{eq:bank_prior}
\B_{c}^{m} \sim p_{c}^{m}(\b x)=\sum\nolimits_{k=1}^{K} \pi_{c}^{m, k} p(\b x \mid y=k),  
\end{equation}
where $\pi_c^{m,k}$ denotes the class prior of class $k$ at the $m$-th sub-bank. We can calculate the class prior via pseudo labels from samples in the sub-bank:
\begin{equation}
\label{eq:pi_mk}
\pi_c^{m,k} = \sum\nolimits_{i=1}^{n_{c,m}}  \mathbbm{1}_{[\operatorname{argmax}(p_i)=k]} /n_{c,m}.
\end{equation}
For the prior regarding proxy label $m$, we have $\bar{\pi}_{c}^{m}=\bar{p}_{c}(\bar{y}=m)$, which can be estimated as $\bar{\pi}_{c}^{m}=n_{c, m} /n_b$. Comparing Eq.~(\ref{eq:bank_prior}) with $\B_c\sim\bar{p}_c(\b x,\bar{y})$, we have the class-conditional density $\bar{p}_c(\b x \mid \bar{y}=m)$ corresponds to the original density $p_c^m(\b x)$, thereby connecting the proxy labels with class labels. In the following, we will design a prior transition function to convert the original classification task into a new sub-bank classification for the local client training.

\subsection{Model Training on Unlabeled Client}
To rely on class priors to supervise client training, we introduce the prior transition function to explicitly connect sub-bank priors and class priors. Denote $g_c$ as the model to perform the sub-bank classification for client $c$. Let $g_c(\b x)_m=\bar{p}_c(\bar{y}=m\mid \b x)$ and $f_c(\b x)_k={p}({y}=k\mid \b x)$, by the Bayes’ rule, for $\forall m \in[K]$:
\begin{equation}
\label{eq:prior_trans}
\begin{aligned}
g_{c}(\b x)_{m} &
=\frac{\bar{p}_{c}(\b x, \bar{y}=m)}{\bar{p}_{c}(\b x)} 
=\frac{\bar{p}_{c}(\b x \mid \bar{y}=m) \bar{p}_{c}(\bar{y}=m)}{\sum_{m=1}^{K} \bar{p}_{c}(\b {x} \mid \bar{y}=m) \bar{p}_{c}(\bar{y}=m)}\\
&=\frac{p_{c}^{m}(\b {x}) \bar{\pi}_{c}^{m}}{\sum_{m=1}^{K} p_{c}^{m}(\b {x}) \bar{\pi}_{c}^{m}}
=\frac{\bar{\pi}_{c}^{m} \sum_{k=1}^{K} \pi_{c}^{m, k} {f_c(\b x)_{k}}/{\pi^{k}}}{\sum_{m=1}^{K} \bar{\pi}_{c}^{m} \sum_{k=1}^{K} \pi_{c}^{m, k} {f_c(\b x)_{k}}/{\pi^{k}}},
\end{aligned}
\end{equation}
where the class prior $\pi^k$ can be estimated by adding up all class priors of sub-banks. By transforming the priors in Eq.~(\ref{eq:prior_trans}) into matrix form, we define the prior transition function as $\mathcal{T}(f_c(\b x^c);\Pi_c, \bar{\b \pi}_c,\b \pi)$, where $\Pi_c \in \R^{K\times K}$ is the matrix of all $\pi_c^{m,k}$, and $\bar{\b \pi}_c=[\bar{\pi}_c^1,\ldots,\bar{\pi}_c^K], \b \pi=[\pi^1,\ldots,\pi^K]$. The prior transition function naturally bridges the class probability and the proxy labels. We could learn the class discrimination knowledge via the sub-bank classification. Given unlabeled client data $\b x^c$ and proxy labels $\bar{y}^c$, we use the prior transition function to perform the 
sub-bank classification with the cross-entropy loss:
\begin{equation}
\label{eq:L_u}
\cL_u = \cL_{ce}(\mathcal{T}(f_c(\b x^c);\Pi_c, \bar{\b \pi}_c,\b \pi), \bar{y}^c).
\end{equation}
Therefore, the overall training loss in Eq.~(\ref{eq:loss_all}) is completed with the new form of $\cL_u$, which is class distribution-aware and subject to each unlabeled client.

\section{Experiments}

\subsection{Dataset and Experiment Setup}
\textbf{Datasets.}
We evaluate our method on two medical image classification tasks: 1) intracranial hemorrhage (ICH) diagnosis for brain CT slices, 2) skin lesion diagnosis for dermoscopy images. For ICH diagnosis, we use the RSNA ICH dataset~\cite{ich} and follow the FedIRM~\cite{fedirm} to randomly sample 25,000 slices which contains 5 subtypes of ICH. For skin lesion diagnosis, we use the HAM10000 dataset~\cite{ham10000}, which contains 10,015 dermoscopy images with 7 skin lesion subtypes. For both datasets, we use 70\% for training, 10\% for validation, and 20\% for test. For data pre-processing, we apply random transformations of rotation, translation, and flipping on 2D images, resize the images into $224\times224$ and normalize them before feeding into an ImageNet~\cite{deng2009imagenet} pre-trained model.
\\
\textbf{Experiment setup.}
We set the number of clients $C$ to 10. For each class of server labeled samples $S_k$, we set $S_k=15$ for ICH diagnosis and $S_k=10$ for skin lesion diagnosis, making the number of labeled data as 75 and 70 respectively. Following previous studies~\cite{li2021model,Wang2020Federated,yurochkin2019bayesian}, we use Dirichlet distribution to simulate the imbalanced classes i.e., sampling $p_k \sim Dir_N(\gamma)$ with $\gamma=1.5$, and allocating a $p_{k,c}$ proportion of instances of class $k$ to client $c$. This yields the proportion of each class in client $c$ in the range $[0.05, 0.5]$. 
We select the best global model from the validation and report the test performance regarding five metrics including AUC, Accuracy, Specificity, Sensitivity, and F1 score over 3 independent runs.
\\
\textbf{Implementation details.}
Following the top-rank solution~\cite{wang2021deep}, we use the DenseNet121~\cite{huang2017densely}. To make a fair comparison, all compared methods are trained on both server and client data, and follow the FedIRM~\cite{fedirm} to apply the warming up for 30 epochs. The thresholds $\tau_\beta$ and $\tau_\alpha$ are set to 0.5 and 0.9. We use the Adam optimizer with momentum terms of $0.9$ and $0.99$, and the batch size is 24. The total communication rounds is 200 with the local training epoch set as 1. 

\subsection{Comparison with State-of-the-art Methods}
\begin{table}[ht]
\centering
\setlength\tabcolsep{1.5pt}
\caption{Results comparison with state-of-the-art methods on two diagnosis tasks.}
\begin{tabular}{@{}lccccc@{}}
\toprule
\multicolumn{6}{c}{Intracranial Hemorrhage Diagnosis} \\ \hline\hline
Methods  & AUC & Accuracy & Specificity    & Sensitivity     & F1           \\ \hline
FedAvg-SL & 88.59\small{$\pm$0.85} & 91.22\small{$\pm$0.17}  & 93.39\small{$\pm$0.09} & 63.64\small{$\pm$0.80} & 59.54\small{$\pm$0.78} \\ \hline
FedIRM~\cite{fedirm} & 60.41\small{$\pm$0.93}  & 72.27\small{$\pm$0.40} & 82.93\small{$\pm$0.29} & 30.13\small{$\pm$0.55}  & 22.93 \small{$\pm$0.31} \\
FedMatch~\cite{fedmatch} & 64.15\small{$\pm$1.76} & 73.09\small{$\pm$0.55} & 84.01\small{$\pm$0.19} & 33.44\small{$\pm$2.60}  & 24.84 \small{$\pm$1.36} \\
FSSL~\cite{semifl_covid} &60.63\small{$\pm$1.92}&72.61\small{$\pm$1.47}  &83.57{$\pm$0.79}   &24.88\small{$\pm$0.39}  &21.98{$\pm$0.93} \\ \hline
FedAvg-FM & 61.54\small{$\pm$1.63} &  74.50\small{$\pm$1.31}  & 86.12\small{$\pm$1.79}& 25.01\small{$\pm$0.55}  &22.55\small{$\pm$0.07} \\ 
FedProx-FM~\cite{fedprox}  & 62.49\small{$\pm$3.41}& 74.95\small{$\pm$3.03}& 85.91\small{$\pm$3.05} & 26.18\small{$\pm$5.24}  & 22.84\small{$\pm$1.55}\\
FedAdam-FM~\cite{fedadam} & 62.42\small{$\pm$3.93}  & 73.85\small{$\pm$0.66} & 85.63\small{$\pm$1.15} & 25.33\small{$\pm$3.62}   & 22.08\small{$\pm$1.02} \\ \hline
imFed-Semi (Ours) & \textbf{82.96\small{$\pm$1.26}} & \textbf{82.56\small{$\pm$0.58}}  & \textbf{90.66\small{$\pm$0.21} } &\textbf{54.58\small{$\pm$3.75}} & \textbf{47.71\small{$\pm$3.59}} \\ \hline
\multicolumn{6}{c}{Skin Lesion Diagnosis}                                                            \\ \hline\hline
Methods  & AUC & Accuracy  & Specificity    & Sensitivity     & F1           \\ \hline
FedAvg-SL & 87.58\small{$\pm$0.28} & 93.32\small{$\pm$0.27} & 89.64\small{$\pm$0.57}& 59.09\small{$\pm$1.34} & 57.09 \small{$\pm$0.68} \\ \hline
FedIRM~\cite{fedirm} & 65.29\small{$\pm$2.26} & 79.05\small{$\pm$1.33} & 90.39\small{$\pm$1.08}& 28.72\small{$\pm$2.22}  & 23.52 \small{$\pm$1.21} \\
FedMatch~\cite{fedmatch}  & 70.90\small{$\pm$1.25} & 84.25\small{$\pm$2.34} & \textbf{93.33\small{$\pm$1.74}} & 29.56\small{$\pm$1.90} & 29.13 \small{$\pm$2.69}\\
FSSL~\cite{semifl_covid} & 70.86{$\pm$1.26}  &83.20{$\pm$1.65}   &93.39{$\pm$0.21}  &28.32{$\pm$0.89}  &27.90{$\pm$1.43}    
\\ \hline
FedAvg-FM & 70.61\small{$\pm$1.79}&  82.67\small{$\pm$0.91} & 91.92\small{$\pm$1.67}& 30.65\small{$\pm$0.91} &29.09 \small{$\pm$0.94}\\
FedProx-FM~\cite{fedprox} & 69.86\small{$\pm$1.48}& 82.01\small{$\pm$1.66} & 91.45\small{$\pm$2.86}& 27.87\small{$\pm$2.69}   & 25.21\small{$\pm$1.14}\\
FedAdam-FM~\cite{fedadam}  & 70.58\small{$\pm$1.86} & 83.22\small{$\pm$2.25} & 92.92\small{$\pm$1.98}& 28.97\small{$\pm$1.87} & 27.85\small{$\pm$1.68} \\ \hline
imFed-Semi (Ours)  & \textbf{77.47\small{$\pm$1.81}}& \textbf{88.94\small{$\pm$1.50}} & 89.81\small{$\pm$2.64} & \textbf{37.48\small{$\pm$2.71}} & \textbf{33.79\small{$\pm$1.75}}        \\ \bottomrule
\end{tabular}
\label{tb:res}
\end{table}
We compare our method with recent state-of-the-art semi-supervised FL methods, including the \textbf{FSSL} (MIA'21)~\cite{semifl_covid} which introduces consistency loss and pseudo labeling into FL, the \textbf{FedIRM} (MICCAI'21)~\cite{fedirm} which enhances the consistency regularization with an inter-client relation matching, and the \textbf{FedMatch} (ICLR'21)~\cite{fedmatch} which applies inter-client consistency and decomposes the model parameter for server and client training. Moreover, we incorporate the widely-used semi-supervised method (FixMatch~\cite{sohn2020fixmatch}, abbreviated as FM) into the baseline FedAvg~\cite{fedavg} and existing FL algorithms for the class imbalance problem, including the \textbf{FedProx}~\cite{fedprox} (MLSys'20) and the \textbf{FedAdam} (ICLR'21)~\cite{fedadam}.
\\\indent Table~\ref{tb:res} lists the results, where FedAvg-SL denotes the supervised learning with full labels on all clients (upper bound). It can be observed that all semi-supervised methods present a great performance drop without sufficient supervision. FedMatch, with the model decomposition on top of the consistency, outperforms other consistency-based semi-supervised FL methods on all metrics. The combination of FM and FL methods shows improvements on the accuracy. Compared with all these methods, our approach achieves the best performance on almost all metrics, with 7.61\% and 4.69\% accuracy boost on both tasks over the second-best. The significant improvement is benefited from our dynamic bank learning scheme, which effectively provides the class distribution-specific supervision to alleviate the model being dominated by majority classes.

\subsection{Analytical Studies on Key Components of the Approach}
\begin{figure}[t!]
\centering
\includegraphics[width=0.99\textwidth]{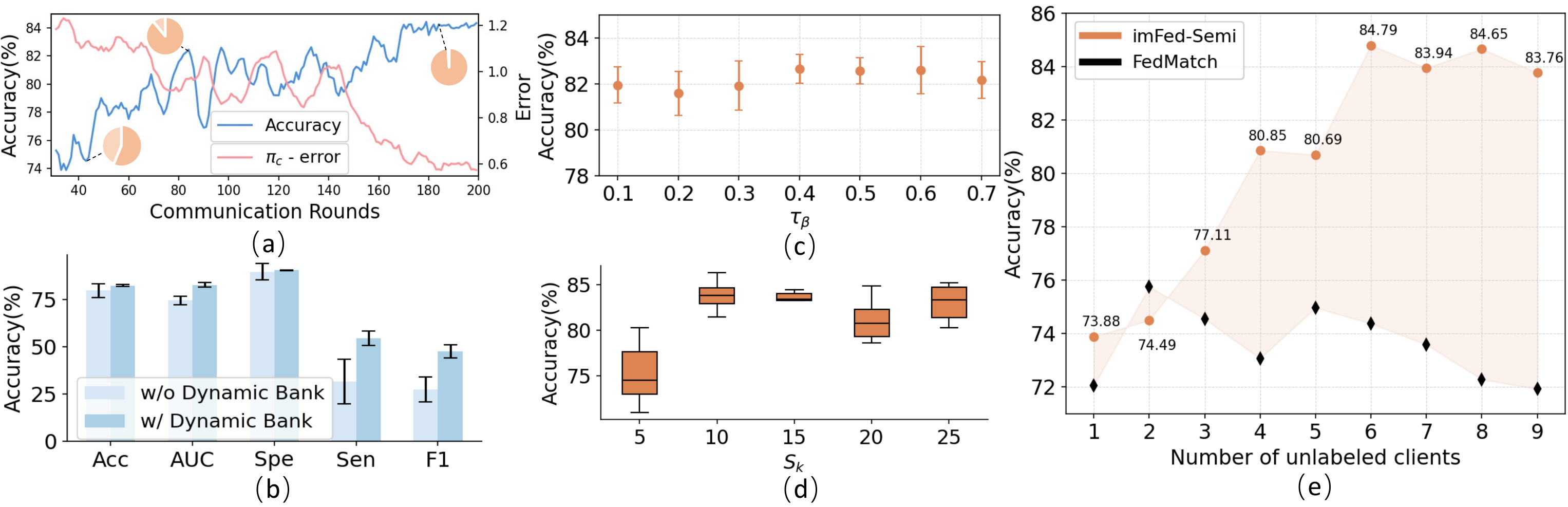}
\caption{Analysis of our approach: (a) learning behaviors of the dynamic bank, (b) ablation on the dynamic bank, (c) effect of different threshold $\tau_\beta$, (d) model performance as increasing server's labeled samples, (e) effect of number of unlabeled clients.}
\label{fig:ablation}
\end{figure}

\textbf{Learning procedure with the dynamic bank.}
We study the learning behavior of our proposed method regarding the dynamic bank construction and class prior estimation on the ICH dataset. As shown in Fig.~\ref{fig:ablation} (a), two curves are the test accuracy and the estimation error (we use the Frobenius norm between the estimated and real class priors), and the pie chart denotes the percentage of data samples collected in the bank. In the beginning, the dynamic bank $\B_c$ gradually collects more confident samples, making the class prior estimation close to the real one and the accuracy is quickly improved (from 74\% to 82\%). With more samples collected, the model takes time to gradually fit different class proportions and the bank finally includes almost all samples (98\%). The local training becomes stable with a high accuracy and low estimation error.
\\
\\
\textbf{Effectiveness of the dynamic bank construction.}
We further investigate the dynamic selection by comparing with a fixed threshold (i.e., 0.9, which is widely adopted). As shown in Fig.~\ref{fig:ablation} (b), the dynamic banks improve the performance on all metrics with smaller standard deviations. Especially, it has 20\% boost on both sensitivity and F1-score. Such improvements benefit from the dynamic way of gathering samples, which helps the model train on diverse classes, thus avoiding fitting to majority classes only. Besides, we study the choice of different thresholds $\tau_\beta$, which can affect the confidence for selecting minority classes. As shown in Fig.~\ref{fig:ablation} (c), the change of $\tau_\beta$ does not affect the test accuracy significantly (less than 1\%), indicating that the way of dynamic construction is the key factor to improve the performance.
\\
\\
\textbf{Studies on scalability with labeled/unlabeled samples.}
We finally analyze the scalability of our method with different numbers of labeled samples at the server and different unlabeled client numbers. The results are shown in Fig.~\ref{fig:ablation} (d) and (e), respectively. For the change of labeled samples, the global model suffers a great performance drop with very few labeled samples, i.e., $S_k=5$. With more samples available, the test accuracy remains stable in a certain range, demonstrating that the performance bottleneck lies in client-side learning. For the scalability of unlabeled clients, the class imbalance issue may be enlarged 
with more clients involved, thereby making client training more difficult. Notably, we find that our method benefits more from the increased unlabeled clients, while the FedMatch might endure a performance drop after introducing more clients.

\section{Conclusion}
We present a new method for the challenging imFed-Semi problem. Our proposed dynamic bank learning scheme provides class proportion-aware supervision for the local client training, significantly improves the global model performance compared with other state-of-the-art semi-supervised learning and FL methods. The effectiveness of our approach is demonstrated on two large-scale real-world medical datasets. The proposed dynamic bank construction is also applicable to other scenarios such as self-supervised learning. In the future, we will further explore the dynamic bank construction by incorporating information from other clients to solve a potential limitation for handling the more severe class imbalance, where each local client may not cover all possible disease categories.
\\
\\
\textbf{Acknowledgement.} This work was supported in part by the Hong Kong Innovation and Technology Fund (Projects No. ITS/238/21 and No. GHP/110/19SZ), in part by the CUHK Shun Hing Institute of Advanced Engineering (project MMT-p5-20), in part by the Shenzhen-HK Collaborative Development Zone, and in part by NSERC Discovery Grant (DGECR-2022-00430).

%
%
%
\bibliographystyle{splncs04}
\bibliography{references}
%




\end{document}